\documentclass[10pt,twocolumn,letterpaper]{article}

\usepackage[pagenumbers]{iccv} 

\usepackage{times}
\usepackage{amsmath}
\usepackage{amssymb}

\definecolor{iccvblue}{rgb}{0.21,0.49,0.74}
\usepackage[pagebackref,breaklinks,colorlinks,allcolors=iccvblue]{hyperref}

\title{Zero-Shot Visual Concept Blending Without Text Guidance}

\author{Hiroya Makino$^{*}$ \qquad Takahiro Yamaguchi \qquad Hiroyuki Sakai\\
Toyota Central R\&D Labs., Inc.}

\begin{document}

\twocolumn[{%
\maketitle
\vspace{-2em}
\begin{center}
    \centering
    \includegraphics[width=0.78\linewidth]{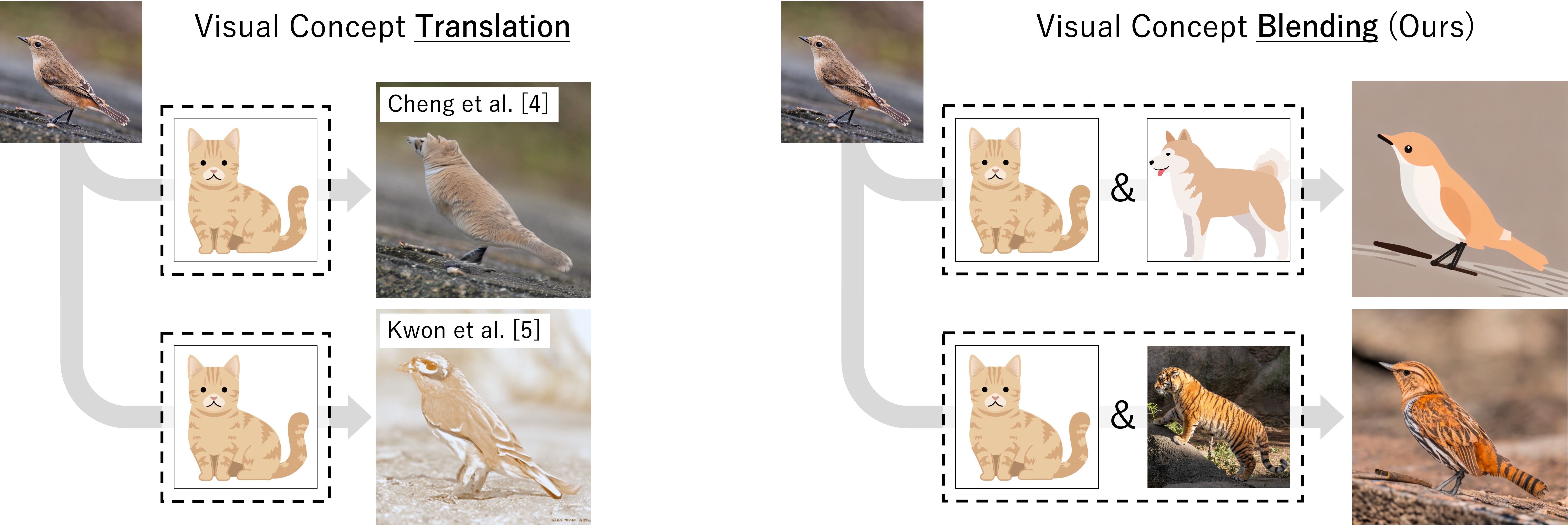}
    \vspace{-0.7em}
    \captionof{figure}{\textbf{Conventional visual concept translation vs. the proposed visual concept blending.}
In conventional methods, it is difficult to distinguish the features that should be transferred because only a single reference image is available.
Our proposed method can effectively control the features we wish to transfer (e.g., illustration style or tiger stripes) by performing operations in the projection space of the CLIP embedding vectors and extracting a common concept from multiple reference images.\label{cover}}
\end{center}
\vspace{1.0em}
}]

\newcommand\blfootnote[1]{%
  \begingroup
  \renewcommand\thefootnote{}\footnote{#1}%
  \addtocounter{footnote}{-1}%
  \endgroup
}
\blfootnote{$*$ Corresponding author. hirom@mosk.tytlabs.co.jp}

\begin{abstract}
We propose a novel, zero-shot image generation technique called ``Visual Concept Blending'' that provides fine-grained control over which features from multiple reference images are transferred to a source image. 
If only a single reference image is available, it is difficult to isolate which specific elements should be transferred.
However, using multiple reference images, the proposed approach distinguishes between common and unique features by selectively incorporating them into a generated output.
By operating within a partially disentangled Contrastive Language-Image Pre-training (CLIP) embedding space (from IP-Adapter), our method enables the flexible transfer of texture, shape, motion, style, and more abstract conceptual transformations without requiring additional training or text prompts. 
We demonstrate its effectiveness across a diverse range of tasks, including style transfer, form metamorphosis, and conceptual transformations, showing how subtle or abstract attributes (e.g., brushstroke style, aerodynamic lines, and dynamism) can be seamlessly combined into a new image. 
In a user study, participants accurately recognized which features were intended to be transferred. 
Its simplicity, flexibility, and high-level control make Visual Concept Blending valuable for creative fields such as art, design, and content creation, where combining specific visual qualities from multiple inspirations is crucial.
Code is available at \url{https://github.com/ToyotaCRDL/Visual-Concept-Blending}.
\end{abstract}

\section{Introduction}
\label{sec:introduction}
\begin{quote} 
\textit{Good artists copy; great artists steal.} -- P. Picasso
\end{quote} 
Picasso\footnote{The adage is often attributed to Pablo Picasso. However, its true origin is a subject of debate.} suggests that merely copying someone else's work is insufficient to create something truly original. 
Instead, great artists take inspiration from existing ideas, make them theirs, and transform them into something new and unique.

In the field of image generation, style transfer is one such example.
The neural style transfer method proposed by Gatys et al.~\cite{gatys2015a, gatys2016} has attracted significant attention because of its ability to preserve the shape and layout of a content image while transferring (stealing) the painting style from a style image.
In recent years, various style transfer techniques have been proposed using generative adversarial networks (GANs) or diffusion models (DMs), and text-based control has become possible.
Although DM-based style transfer with text input enables high-resolution and flexible style transfer in real-world artworks and designs, the source of a style may not always be described precisely in the text.
As indicated by Parmar et al.~\cite{parmar2023}, images contain far more varied and extensive information than text, making it extremely difficult to convert the texture, abstract composition, or atmosphere of a reference image into text.

To transfer visual concepts effectively while preserving the shape and layout of a content image, Cheng et al.~\cite{cheng2023} and Kwon et al.~\cite{kwon2023} proposed visual concept translation (VCT).
VCT transfers the features of a single reference image onto another image.
By taking an image directly as input rather than text, VCT enables the transfer of more complex concepts.

However, VCT, which employs images rather than text to represent concepts, faces two primary challenges.
First, it has no control over which elements of the reference image are transferred.
For example, as shown in Fig.~\ref{cover}~(left), if the reference image is an illustration of an orange tabby cat, it is impossible to specify whether the transferred features should be the cat's stripes, the overall feline shape or the illustration style.
Second, it is challenging to transfer higher-order concepts.
Existing methods rely on the loss function of a DM.
While these methods can alter appearance, handling more abstract concepts (e.g., ``flying'' or ``degenerated'') remains challenging.

To address these challenges, this study proposes \textit{Visual Concept Blending}.
Our method works within linearly transformed Contrastive Language-Image Pre-training (CLIP)~\cite{radford2021} embedding space from IP-Adapter~\cite{ye2023}, a model built on Stable Diffusion, and extracts the common or distinct elements between two images.
It then transfers these features onto another image to generate the final output (see Fig.~\ref{cover}~(right)).
Using two reference images instead of one, we can determine which features of the reference images should be transferred.
Moreover, by operating within the CLIP embedding space, we can achieve more flexible transformations, including ``abstract shape changes'' or ``motion,'' rather than simply transferring colors or textures.
The method is extremely simple and robust because it operates solely within the embedding space of a pre-trained model and requires no domain-specific fine-tuning.

The remainder of the paper is organized as follows.
Section~\ref{sec:related_work} provides an overview of related work ranging from early style transfer to VCT.
Section~\ref{sec:method} describes the proposed method in detail.
In Section~\ref{sec:experiments}, we present the various results produced from different combinations of source and reference images and demonstrate the effectiveness of our approach through a user study.
Section~\ref{sec:limitations} discusses the limitations of the proposed approach.
Finally, Section~\ref{sec:conclusion} concludes the study.

%
%
\section{Related Work}
\label{sec:related_work}
In 2015, Gatys et al.~\cite{gatys2015a, gatys2016} introduced style transfer using neural networks. 
Style transfer refers to extracting style information from a reference image and applying it to another image (in this study, it is called a source image).
Gatys et al. used intermediate features from VGG-19~\cite{simonyan2015} (pre-trained on ImageNet~\cite{russakovsky2015}) to generate images that preserve the structure of the source while incorporating the style of the reference.

However, this approach requires optimization for each style, making the process time-consuming.
To address this, some methods pre-trained networks to convert images into specific styles~\cite{johnson2016a, ulyanov2016, ulyanov2017, dumoulin2017, li2018}.
However, they cannot handle unseen styles.
To solve this issue, techniques for fast arbitrary-style transfers have been introduced~\cite{chen2016, huang2017, li2017, sheng2018}.

In addition to the feedforward neural network approaches mentioned above, several style transfer methods using GANs~\cite{goodfellow2014} have also been proposed.
Isola et al.~\cite{isola2017} introduced pix2pix, a supervised learning method that relies on paired image datasets for style transfers.
However, preparing paired images is expensive.
To overcome this problem, Zhu et al.~\cite{zhu2017}, Yi et al.~\cite{yi2017}, and Park et al.~\cite{park2020} proposed unsupervised learning methods that use unpaired datasets.
Choi et al.~\cite{choi2018} extended CycleGAN~\cite{zhu2017} to handle multiple styles or attribute domains in a single model.
However, both supervised and unsupervised methods typically require large amounts of training data.
Contrarily, some techniques allow style transfer learning using only a small amount of data for new tasks~\cite{liu2019b, lin2020b, lin2020, lin2021}.

Recently, image transformation using DMs has become popular because they can generate higher-resolution images than GANs~\cite{dhariwal2021}. 
Meng et al.~\cite{meng2022} proposed adding noise to an input image and subsequently denoising it to produce realistic images.
Su et al.~\cite{su2023} introduced a technique that separates the DM for the input domain from that for the output domain, thereby enabling style transfer between domains where paired data are difficult to collect.
In addition, Li et al.~\cite{li2023b} proposed a DM that uses a Brownian bridge as a stochastic process, enabling style transfer even between domains with low similarity.

Style transfer with DMs has mainly focused on relatively simple styles, such as those used in dataset labels (e.g., ``animal: tiger, hummingbird,'' ``colorization,'' and ``semantic segmentation'').
More complex image transformations using text-to-image DMs have also been explored recently.
Hertz et al.~\cite{hertz2022} introduced a complex feature-based image transformation method by editing text prompts.
This method assumes that the input image is represented as text. 
However, methods such as Imagic~\cite{kawar2023} and pix2pix-zero~\cite{parmar2023} enable image transformation without relying on the text representation of the input image.

Image transformation with text-to-image DMs has enabled more complex style transfer.
However, the transformation style should also be specified in the text.
Recalling the initial definition in this section, style transfer refers to the transfer of style information from a reference image to a source image.
If all the style information from a reference image can be fully captured in the text, style transfer with text-to-image DM is feasible.
However, as noted by Parmar et al.~\cite{parmar2023}, images contain far more information than text, making it difficult to capture them purely in textual form.

Contrary to transformations that use text-to-image DMs, Cheng et al.~\cite{cheng2023}, Kwon et al.~\cite{kwon2023}, Wang et al.~\cite{wang2023b}, and Zhang et al.~\cite{zhang2023d} proposed methods that generate new images by combining the content of a source image with the style of a single reference image.
However, because the style originates from a single reference image, it is difficult to control the specific features that are transferred. 
Additionally, because these methods rely on a DM's loss function, they can modify their appearance but struggle to transfer higher-level concepts.

Conversely, our proposed method uses two reference images, which allows for extracting common or distinct features.
This approach enables explicit control of the transferred elements.
Furthermore, by operating in the projection space of CLIP embeddings, our method can transfer higher-level concepts encoded in CLIP without any additional training.

%
%
\section{Visual Concept Blending}
\label{sec:method}
\subsection{Preliminaries}
\subsubsection{CLIP}
\label{sec:clip}
CLIP~\cite{radford2021} encodes images and texts as embedding vectors.
It consists of two main components: an image and a text encoder. 
CLIP is trained on a large dataset of image-text pairs and learns embeddings, where pairs with similar meanings have a high cosine similarity.

Recently, several studies have reported that projecting CLIP embeddings onto another space can produce partially disentangled embeddings.
Moayeri et al.~\cite{moayeri2023} demonstrated that a linear projection of image embeddings, rather than text embeddings, results in partial disentanglement.
Similarly, Materzynska et al.~\cite{materzynska2022} demonstrated that a linear projection of CLIP embeddings can separate linguistic features from visual features in an image.
Furthermore, Bhalla et al.~\cite{bhalla2024} revealed that CLIP image embeddings can be linearly transformed into vectors, in which each dimension has an interpretable semantic meaning.
These findings suggest that each dimension of the projected embedding vectors can independently capture distinct semantic features.

\subsubsection{IP-Adapter}
IP-Adapter~\cite{ye2023} is an extension of Stable Diffusion that can take image features as an additional condition in text-to-image generation. 
In text-to-image diffusion models, the diffusion process gradually adds Gaussian noise to the training data over a fixed Markov chain of $T$ steps, and the denoising process eliminates noise to generate samples from random noise.
Typically, the training objective of a diffusion model, denoted as $\boldsymbol{\epsilon}_\theta(\boldsymbol{x}_t, \boldsymbol{c}, t)$, is defined as:
\begin{equation}
L(\theta) \;=\; \mathbb{E}_{\boldsymbol{x}_0,\boldsymbol{\epsilon}, \boldsymbol{c}, t}
  \bigl\|\boldsymbol{\epsilon}
        \;-\;
         \boldsymbol{\epsilon}_\theta(\boldsymbol{x}_t, \boldsymbol{c}, t)
  \bigr\|^2,
\end{equation}
where $\boldsymbol{x}_0$ is the real data with condition $\boldsymbol{c}$, $t$ is the diffusion timestep, and $\boldsymbol{x}_t$ denotes noisy data at step $t$.
IP-Adapter is based on an open-source Stable Diffusion~\cite{rombach2022a}, which uses a UNet~\cite{ronneberger2015a} with cross-attention layers to condition on text.
Given text embeddings $\boldsymbol{c}_\mathrm{txt}$ and query features $\mathbf{Z}$, the cross-attention layer output is defined as:
\begin{align}
\label{eq:cross_attention_text}
    \mathbf{Z}' = \mathrm{Attention}(\mathbf{Q}, \mathbf{K}, \mathbf{V}) = \mathrm{Softmax}\left(\frac{\mathbf{Q} \mathbf{K}^\top}{\sqrt{d_k}}\right) \mathbf{V},
\end{align}
where $\mathbf{Q} = \mathbf{Z}\mathbf{W}_q$ is the query matrix, $\mathbf{K} = \boldsymbol{c}_\mathrm{txt}\mathbf{W}_k$ is the key matrix, $\mathbf{V} = \boldsymbol{c}_\mathrm{txt}\mathbf{W}_v$ denotes the value matrix, and $d_k$ is the key dimension.
$\mathbf{W}_q$, $\mathbf{W}_k$, and $\mathbf{W}_v$ are weight matrices of the linear projection layers.

To incorporate image features, IP-Adapter introduces decoupled cross-attention layers with separate key and value matrices:
\begin{align}
\label{eq:cross_attention_image}
    \mathbf{Z}'' = \mathrm{Attention}(\mathbf{Q}, \mathbf{K}', \mathbf{V}') = \mathrm{Softmax}\left(\frac{\mathbf{Q} (\mathbf{K}')^\top}{\sqrt{d_k}}\right) \mathbf{V}',
\end{align}
where $\mathbf{Q} = \mathbf{Z}\mathbf{W}_q$ , $\mathbf{K}'=\boldsymbol{c}_\mathrm{img}\mathbf{W}'_k$ , and $\mathbf{V}'=\boldsymbol{c}_\mathrm{img}\mathbf{W}'_v$.
The final output is the sum:
\begin{align}
\label{eq:cross_attention_final}
    &\mathbf{Z}^\mathrm{new} = \mathbf{Z}' + \mathbf{Z}''\\
    & \quad = \mathrm{Softmax}\left(\frac{\mathbf{Q} \mathbf{K}^\top}{\sqrt{d_k}}\right) \mathbf{V} + \mathrm{Softmax}\left(\frac{\mathbf{Q} (\mathbf{K}')^\top}{\sqrt{d_k}}\right) \mathbf{V}'.
\end{align}
Practically, the original UNet parameters (including $\mathbf{W}_q$) remain frozen, and only the new projection layers $\mathbf{W}'_k$ and $\mathbf{W}'_v$ are trained.

In this study, we assumed that the CLIP embedding space in IP-Adapter is partially disentangled, similar to the findings discussed in Section~\ref{sec:clip}. 
Because IP-Adapter linearly transforms image embeddings before feeding them into the cross-attention layers, it likely decomposes these embeddings so that some semantic features can be extracted and combined.
Under this assumption, we explored the semantic dimensions of the IP-Adapter embedding space and examined how manipulating these dimensions influences the generated images.

Furthermore, because IP-Adapter adds only attention layers without altering the base UNet model, it remains compatible with other controllable adapters such as ControlNet~\cite{zhang2023e}.
In this study, we provided depth maps~\cite{bhat2023} for input images and used ControlNet to generate images consistent with these maps.
We introduced parameter $d (\geq0)$ to control the extent to which the structure of the depth map is preserved.
When $d=0$, the depth map is ignored. 
As $d$ increases, the generated image follows the depth map structure more closely.

\begin{figure}[!t]
  \centering
  \includegraphics[width=0.99\linewidth]{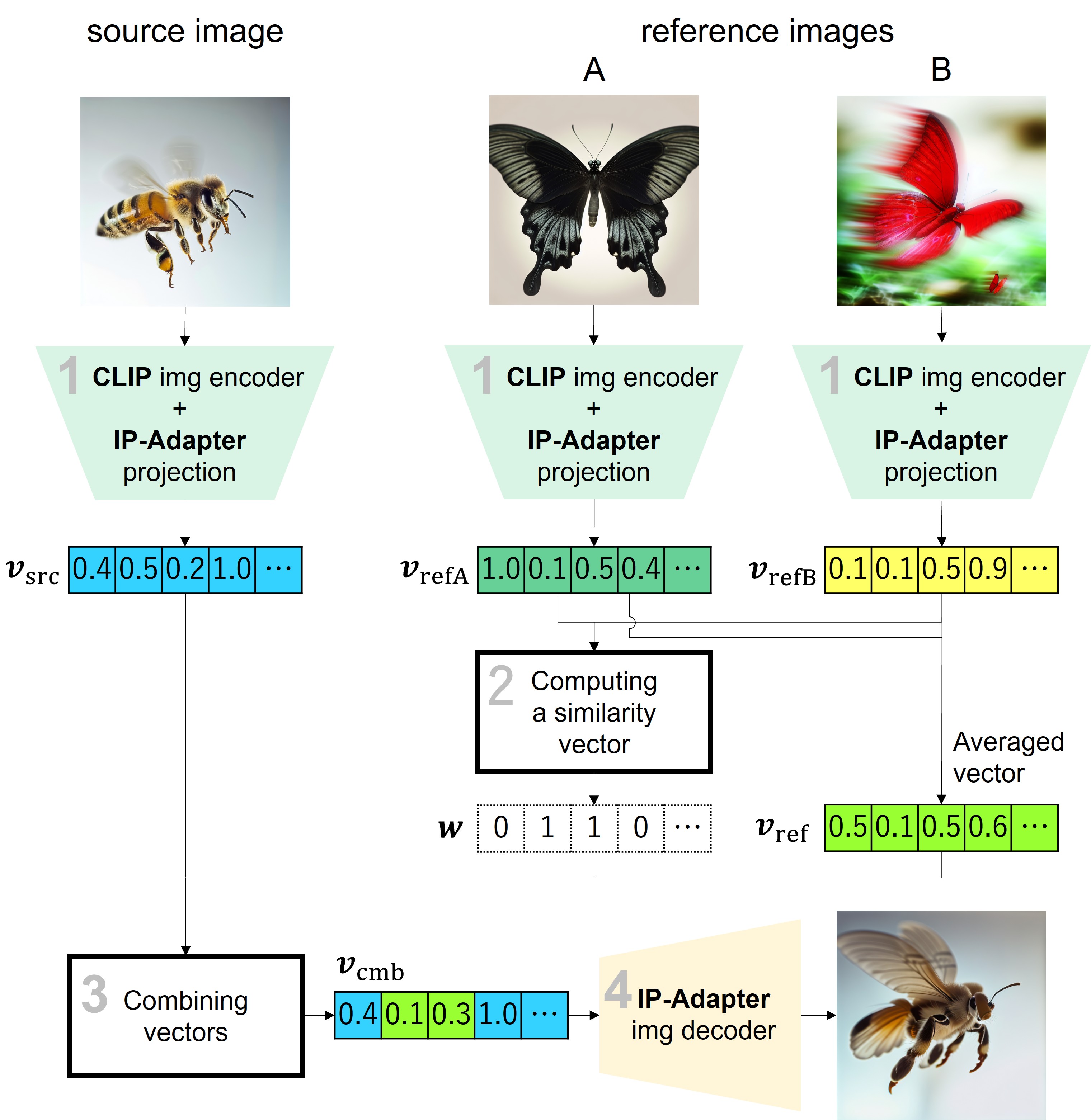}
  \vspace{-0.7em}
  \caption{\textbf{Architecture of the proposed method.}}
  \label{architecture}
\end{figure}

\subsection{Overall Framework}
\label{sec:framework}
We propose a method to generate new images by transferring common or distinct features from multiple reference images to a source image.
We perform feature transfer through the operations in the IP-Adapter's CLIP embedding space.
The proposed method comprises four main steps:
\begin{enumerate}
  \item \textbf{Encoding Images}: Convert the source and reference images into embedding vectors.
  \item \textbf{Computing a Similarity Vector}: From the reference embeddings, compute a similarity vector to extract common or distinct features.
  \item \textbf{Combining Source and Reference Vectors}: Use the similarity vector to combine the source and reference embeddings.
  \item \textbf{Generating Images}: Generate an image from the combined embedding.
\end{enumerate}
The architecture for transferring common features is illustrated in Fig.~\ref{architecture}. 
The numbers in the figure correspond to the aforementioned steps.

To encode images, we used the IP-Adapter's CLIP image encoder.
Let $\boldsymbol{v}_\mathrm{src}$ be the embedding of the source image and $\boldsymbol{v}_\mathrm{refA}$ and $\boldsymbol{v}_\mathrm{refB}$ be those of the two reference images.

If the IP-Adapter's CLIP embedding space is disentangled, common features between reference images should be embedded in the same dimensions. 
Thus, the difference in these dimensions between $\boldsymbol{v}_\mathrm{refA}$ and $\boldsymbol{v}_\mathrm{refB}$ is small. 
In our method, we identify common features by checking whether the absolute difference between $\boldsymbol{v}_\mathrm{refA}$ and $\boldsymbol{v}_\mathrm{refB}$ in each dimension is below threshold $\theta$.
To formalize this, we define a similarity vector $\boldsymbol{w}$. 
This vector shares the same dimensionality as the IP-Adapter's CLIP embedding space: dimensions representing common features are $1$, and others are $0$.
Each dimension $i$ of $\boldsymbol{w}$ is computed as follows:
\begin{align}
  w_i = 
  \left\{
    \begin{array}{ll}
      1 & \mathrm{if} \, \, |v_{\mathrm{refA},i}-v_{\mathrm{refB},i}| < \theta \\
      0 & \mathrm{otherwise}
    \end{array}
    \right.
\end{align}

In the vector combination step, we merge the source vector $\boldsymbol{v}_\mathrm{base}$ with reference vectors $\boldsymbol{v}_\mathrm{refA}$ and $\boldsymbol{v}_\mathrm{refB}$.
Because there are two reference images, we first compute their average to form a single reference vector: $\boldsymbol{v}_\mathrm{ref} = (\boldsymbol{v}_\mathrm{refA} + \boldsymbol{v}_\mathrm{refB})/2$.
To extract common features, we integrate the shared attributes from the reference vector $\boldsymbol{v}_\mathrm{ref}$ while preserving the remainder from the source image vector $\boldsymbol{v}_\mathrm{base}$. 
This is formulated as:
\begin{align}
  \boldsymbol{v}_\mathrm{cmb} = (\boldsymbol{1} - \boldsymbol{w}) \odot \boldsymbol{v}_\mathrm{base} + \boldsymbol{w} \odot \boldsymbol{v}_\mathrm{ref},
\end{align}
where $\odot$ denotes the Hadamard (element-wise) product. 
By reversing the coefficients, distinct features can be transferred from the reference images (specifically those in A but not in B) onto the source image.
\begin{align}
  \boldsymbol{v}_\mathrm{cmb} = \boldsymbol{w} \odot \boldsymbol{v}_\mathrm{base} + (\boldsymbol{1} - \boldsymbol{w}) \odot \boldsymbol{v}_\mathrm{refA}.
\end{align}

Finally, we used the IP-Adapter's decoder to generate an image from the combined vector $\boldsymbol{v}_\mathrm{cmb}$.

\begin{figure}[!t]
  \centering
  \includegraphics[width=0.99\linewidth]{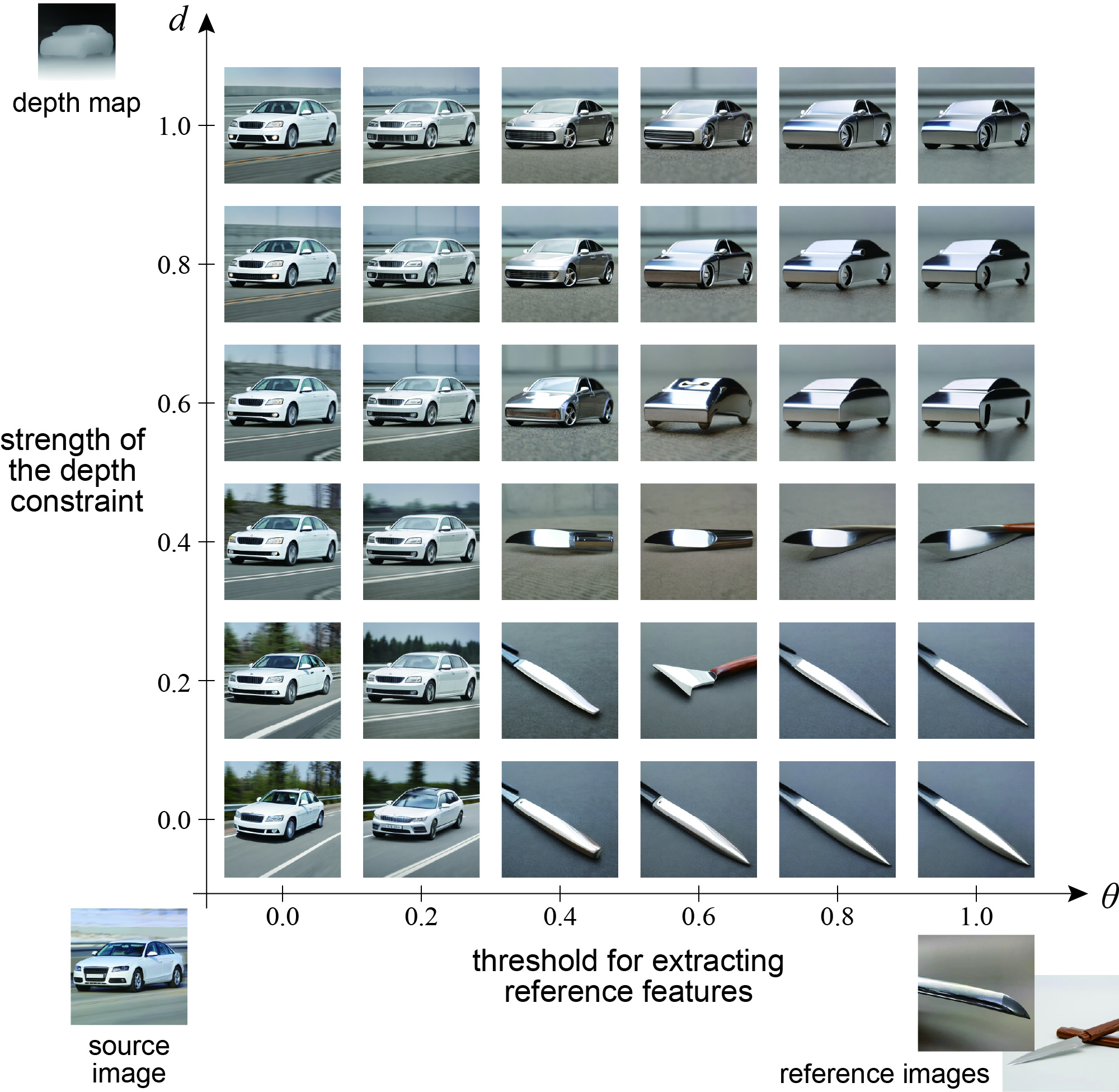}
  \vspace{-0.7em}
  \caption{\textbf{Parameter sensitivity analysis.} Here, $\theta$ is the threshold for extracting reference features, and $d$ is the strength of the depth constraint.}
  \label{batch1}
\end{figure}

\paragraph*{Parameters.}
Our method has two main parameters, which strongly affect the generated results.
Fig.~\ref{batch1} presents the results of the parameter sensitivity analysis.
In this example, the source image was a car, whereas the reference images were a knife and a sword.
The two key parameters of the method are reference threshold $\theta$ and depth constraint strength $d$. 
The generated image can change significantly depending on the parameter values.
For example, at $(\theta, d) = (0.2, 0.6)$, a horizontal line appeared on the front bumper; however, at $(\theta, d) = (0.0, 0.6)$, it did not appear. 
This suggests that the sharpness of the knife and sword was transferred.
Moreover, at $(\theta, d) = (0.4, 1.0)$, the image adopted a metallic sporty design, whereas at $(\theta, d) = (0.8, 0.8)$, it resembled a conceptual car.
Naturally, suitable parameter values depend on images and specific context (for instance, whether designing an early conceptual car or making final design adjustments).
Additional analyses are presented in the Appendix.

%
%
\section{Experiments}
\label{sec:experiments}
\subsection{Qualitative Results}

\begin{figure*}[!t]
  \centering
  \includegraphics[scale=0.195]{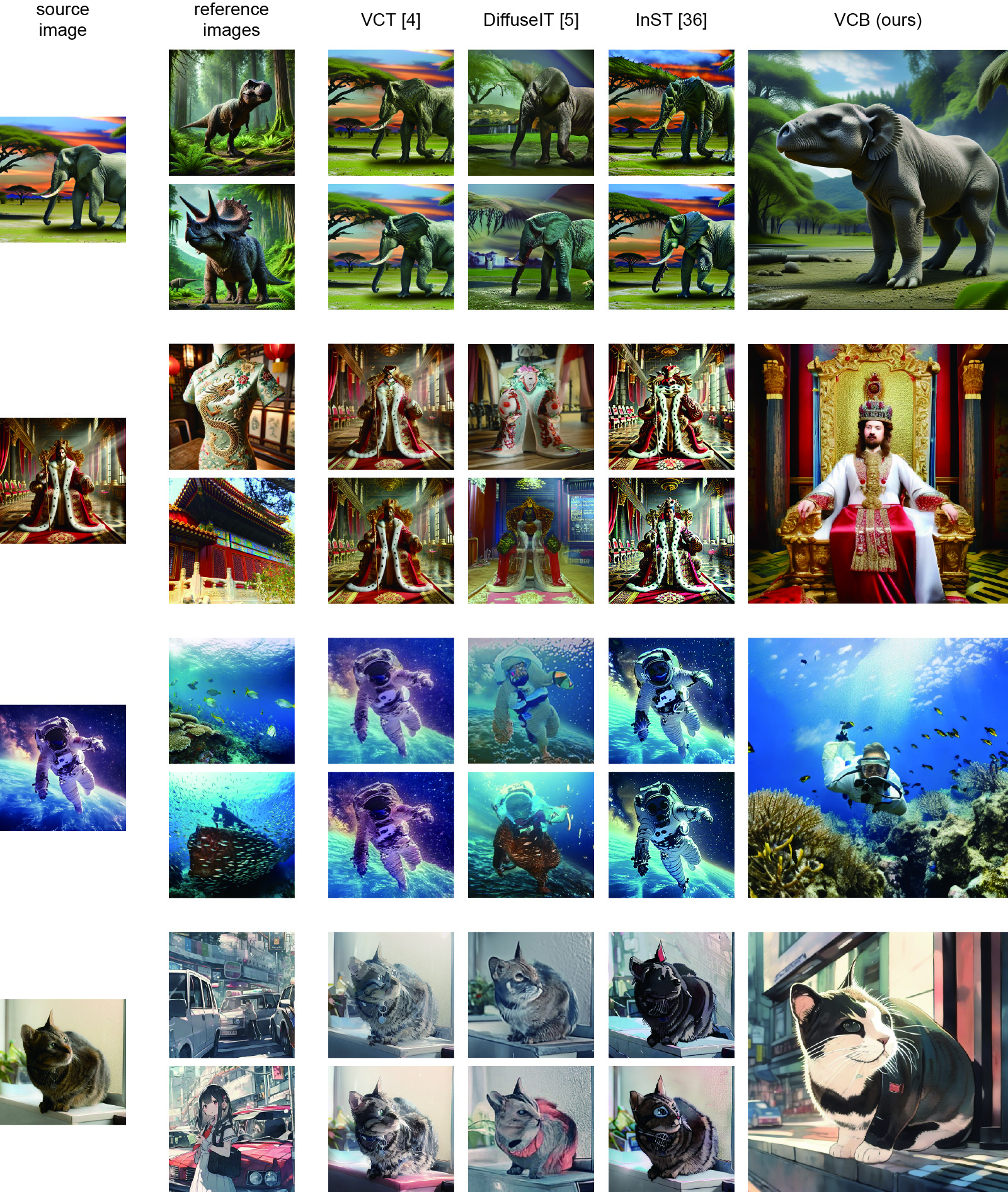}
  \vspace{-0.7em}
  \caption{\textbf{Blending the common features of reference images and comparison with existing single-reference methods.} For VCB, $\theta=0.4$ and $d=0.0, 0.0, 0.1, 0.6$, respectively.}
  \label{qualitative_results_1_1}
\end{figure*}

\begin{figure}[!t]
  \centering
  \includegraphics[scale=0.195]{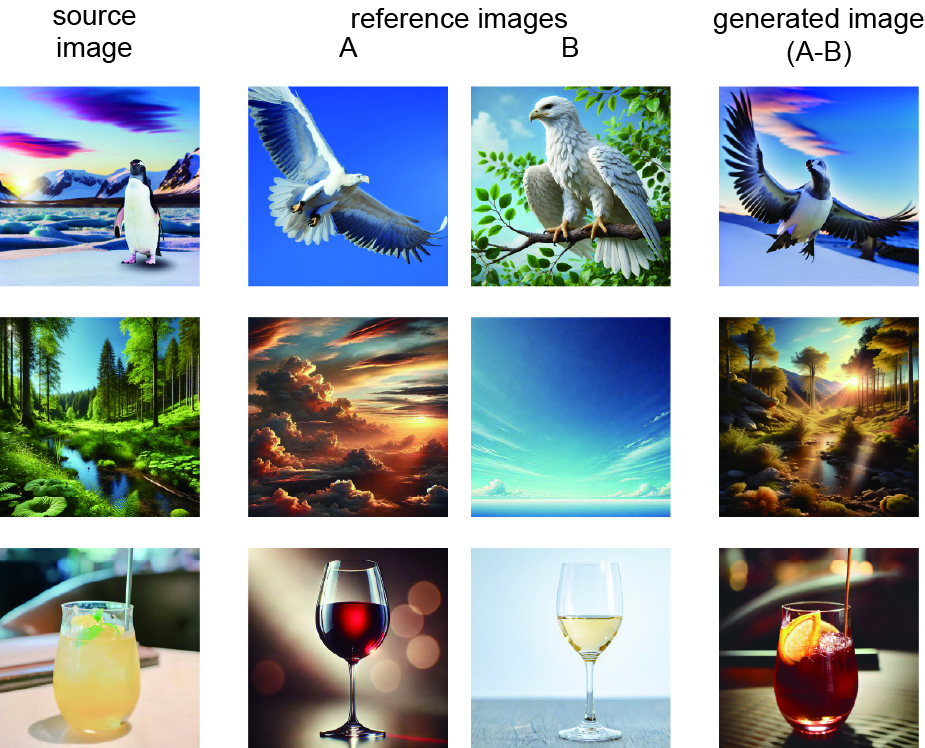}
  \vspace{-0.7em}
  \caption{\textbf{Blending the distinct features between reference images.} $\theta=0.6$ and $d=0.0, 0.0, 1.0$, respectively.}
  \label{qualitative_results_2_1}
\end{figure}

Fig.~\ref{qualitative_results_1_1} shows the results of transferring features from the reference images and compares them with those of existing methods (VCT~\cite{cheng2023}, DiffuseIT~\cite{kwon2023}, and InST~\cite{zhang2023d}).
The proposed method extracts features common to two reference images to generate a new image.
Existing methods independently extract features from each image to generate a new image.

The first row of Fig.~\ref{qualitative_results_1_1} shows an example in which dinosaur traits were transferred to elephants walking in grasslands.
In the image generated using our method, the elephant's body was slimmer, the trunk and ears were smaller, and the face was altered, reflecting a more ancestral form.
Because the reference images shared similar backgrounds, the generated backgrounds reflected the shared features.
However, although existing methods changed the surface of an elephant's body, they failed to adequately incorporate the intended dinosaur traits.

The second row of Fig.~\ref{qualitative_results_1_1} shows an example in which the shared elements of a ``Chinese Dress'' and an ``Asian house'' were transferred onto a European king.
The shared feature of these two reference images was the Asian culture.
In the image generated by our proposed method, both the European king and surrounding atmosphere exhibited an Asian-like transformation.
However, in existing methods, either the European atmosphere remains unchanged, or the style transformation is so unnatural that the king's face is lost.

The third row of Fig.~\ref{qualitative_results_1_1} shows an example of transferring a ``sea-like'' quality onto an astronaut.
After applying the proposed method, the astronaut dived into the ocean and transformed into a scuba diver, while the background changed from outer space to an underwater environment.
Although the existing methods produced some color changes, they failed to transfer the intended shared elements.

The fourth row of Fig.~\ref{qualitative_results_1_1} shows an example of transferring an ``anime-like'' style to an image of a cat.
Both reference images were complex illustrations, making it difficult to extract specific features from either image alone.
However, by providing two reference images, the proposed method successfully extracted the desired style.

Fig.~\ref{qualitative_results_2_1} shows how the features in reference image A, but \textit{not} in reference image B, were transferred onto the source image.
The first row of Fig.~\ref{qualitative_results_2_1} demonstrates the transfer of the features found in a flying white eagle, but not in a stationary one, onto a penguin.
The key difference is that the bird was in flight; therefore, in the generated image, the penguin had wings and a flight posture.
Meanwhile, both reference images were of white eagles; therefore, color was a shared feature.
Consequently, color was not identified as a different feature, and the penguin in the generated image retained its original color.
Similarly, the second row of Fig.~\ref{qualitative_results_2_1} shows a forest with a small stream changing into a sunset scene. The third row shows a refreshed fruit cocktail turning into a red wine-based cocktail.

In summary, using the existing visual concept transfer approaches, our method can extract essential features that are difficult to isolate.
Moreover, it allows the expression of complex features that are difficult to capture using typical text-to-image DMs (indeed, who could describe the anime-like concept in Fig.~\ref{qualitative_results_1_1} using only the text?).
For other comparisons, please refer to the Appendix.

\subsection{User Study}
\subsubsection{Overview}
This user study aimed to verify whether the proposed method can transfer features in the intended direction.

\begin{figure}[!t]
  \centering
  \includegraphics[width=0.72\linewidth]{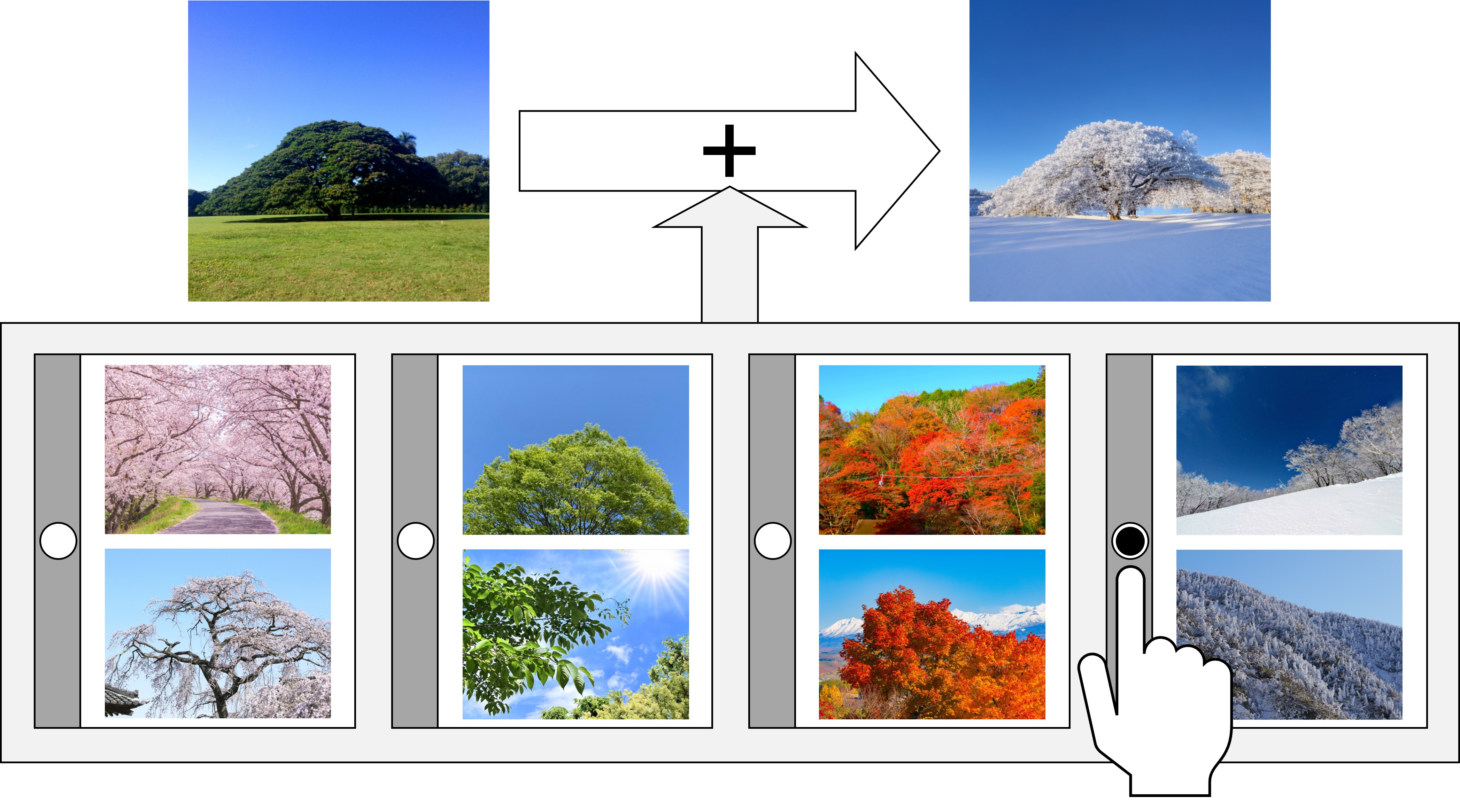}
  \vspace{-0.7em}
  \caption{\textbf{Example of the user study.} The figure shows one source image (upper left), four pairs of reference images (bottom), and one generated image (upper right). In this example, the generated image has snow on the tree leaves and ground; hence, the fourth pair is the correct answer.}
  \label{user_study_example}
\end{figure}

For each question, one source image, four pairs of reference images, and one generated image were presented.
The generated image was typically created using one of the four reference pairs. 
If the participants correctly selected the pair used to generate the image, the features were transferred as intended.
For example, in Fig.~\ref{user_study_example}, the generated image shows snow on the tree leaves and ground; hence, the fourth pair (winter) is the correct answer.

However, confounding biases should be considered when considering the probability of choosing the correct pair. 
If the source image already had many features of the correct pair, even without feature transfer, the participants were more likely to pick that pair.
In this experiment, we prepared a baseline image generated solely from the source image without any reference features.
If the probability of choosing the correct pair with the reference features is higher than that of the baseline, we can conclude that the features are transferred as intended.

We now describe how to compare these probabilities while accounting for confounding biases. 
Let $I_\mathrm{gen} = G(I_\mathrm{src}, R_\mathrm{tgt})$ be a function that transfers the common features of the correct pair $R_\mathrm{tgt}$ to the source image $I_\mathrm{src}$ to produce $I_\mathrm{gen}$.
Moreover, let $P(R_\mathrm{tgt} \mid I_\mathrm{gen})$ be the probability that the subject selects the correct pair $R_\mathrm{tgt}$ when presented with $I_\mathrm{gen}$. 
If
\begin{align}
P\bigl(R_\mathrm{tgt} \mid G(I_\mathrm{src}, R_\mathrm{tgt})\bigr) > P\bigl(R_\mathrm{tgt} \mid G(I_\mathrm{src}, \emptyset)\bigr),
\label{eq_compare}
\end{align}
then feature transfer has been achieved. 
Here, $G(I_\mathrm{src}, \emptyset)$ represents an image generated solely from the source image, with no reference features.

In this user study, we prepared 15 questions for each of three categories: artwork, car, and interior.
Each category had three source images and four pairs of reference images, producing 15 questions per category (including the baseline).
A total of 100 participants participated in this study.
This user study was approved by our Institutional Review Board, and each participant provided informed consent before participating.
For further ethical and societal considerations, please refer to the Appendix.

\subsubsection{Results \& Discussion}
\begin{figure*}[!t]
  \centering
  \includegraphics[width=0.86\linewidth]{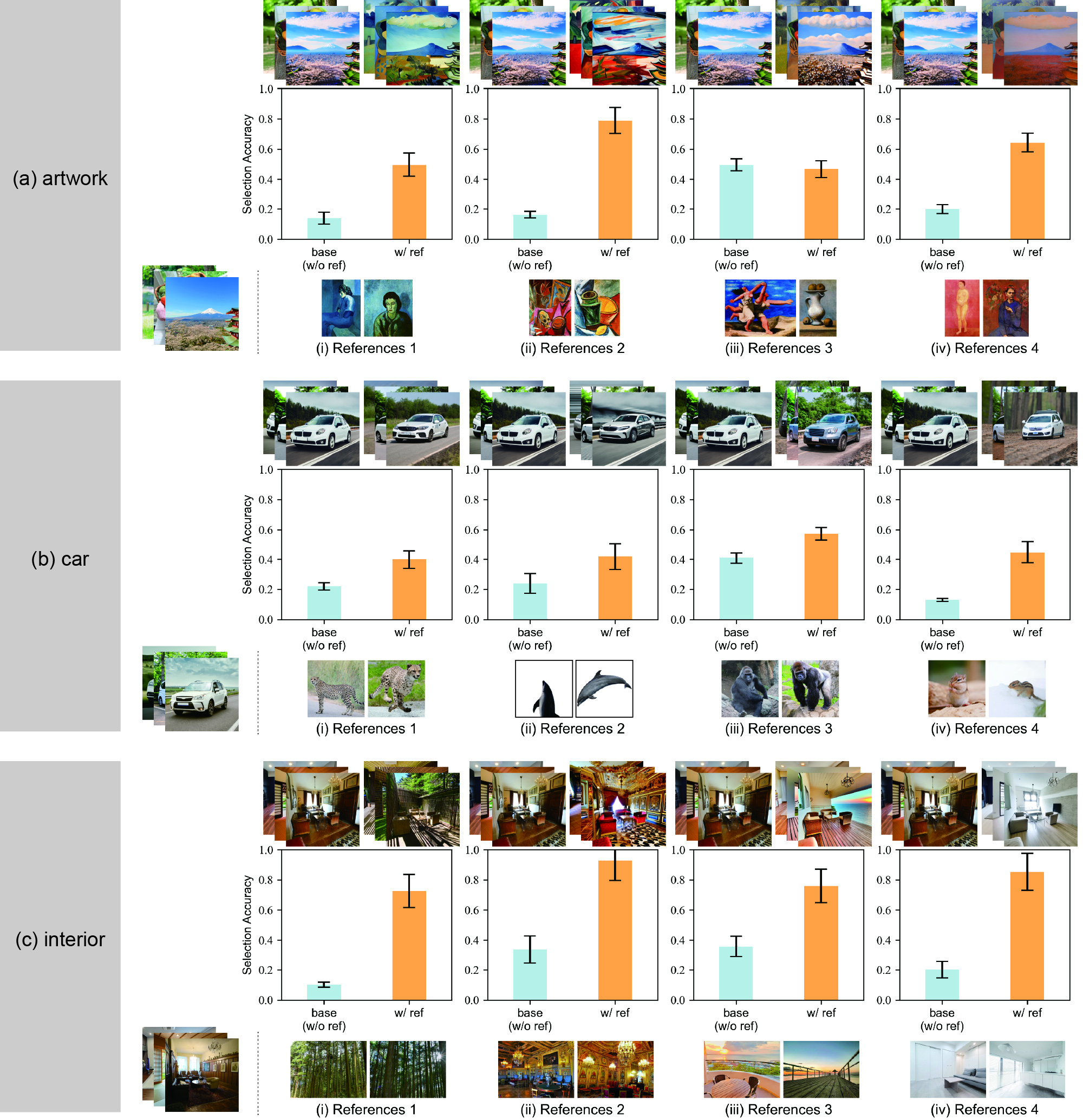}
  \vspace{-0.7em}
  \caption{\textbf{Results of the user study.} In each category, the source image is on the left, the reference image pairs are at the bottom, and the generated images are at the top.  
  Participants were shown a generated image alongside the source image and four reference pairs labeled (i)--(iv).  
  The fraction of participants who correctly identified the actual reference pair is \textit{selection accuracy} (``w/ ref,'' on the right of each graph).  
  As a baseline, we also show \textit{selection accuracy} when no reference features were used (``w/o ref,'' on the left).  
  Artwork references are from WikiArt (\url{https://www.wikiart.org/en/pablo-picasso/}).}
  \label{user_study}
\end{figure*}

Fig.~\ref{user_study} presents the results of the user study.
We recorded the probability (accuracy) of the participants selecting the correct reference pair in each category and for each reference pair, under two conditions: a baseline (source only) and reference condition (using the correct pair $R_\mathrm{tgt}$).
In most cases, in all categories, the accuracy $P\bigl(R_\mathrm{tgt} \mid G(I_\mathrm{src}, R_\mathrm{tgt})\bigr)$ under the reference condition exceeded baseline $P\bigl(R_\mathrm{tgt} \mid G(I_\mathrm{src}, \emptyset)\bigr)$.
This suggests that adding the reference pair's distinctive features to the generated image helps participants more accurately identify ``which pair was used.''

In the Artwork category (Fig.~\ref{user_study}(a)), we successfully transferred the intended features for three of the four reference pairs.
All the four reference images in this category were obtained from Picasso.
Picasso's style evolved over several periods: (i) the Blue Period, (ii) the African Period (Cubism), (iii) the Neoclassical and Surrealist Periods, and (iv) the Rose Period.
Among these, (i), (ii), and (iv) feature distinctive color usage, and (ii) presents a unique brushstroke style, allowing us to successfully transfer these concepts to the source image.
However, for (iii), although the generated image partially reflected the features of the reference image, its colors were close to real-world hues, making the probability of choosing the correct pair almost similar to that of the baseline image.

In the Car category (Fig.~\ref{user_study}(b)), we successfully transferred the intended features for all four reference pairs.
The four references in this category were cheetahs, dolphins, gorillas, and squirrels.
We chose these references such that the cheetah would convey agility; the dolphin, a smooth, streamlined form; the gorilla, a robust feel; and the squirrel, a small, cute impression.
Although cars and animals might appear unrelated, designs that mimic biological structures are widely known as biomimetics.
In this study, we adopted a biomimetics-like approach using only images to transfer the external features of these animals to the source image.

For the interior category (Fig.~\ref{user_study}(c)), we transferred the intended features for all four reference pairs.
When designing interiors, it is common to draw inspiration from external images to set the atmosphere and concepts of a space.
In this experiment, we used four concepts as references (forest, palace, resort, and minimal design) and transferred their characteristics to the source image.
The forest reference emphasized a natural color palette of wood grains and greenery, creating a calm interior atmosphere.
The palace references transferred colors and patterns that were evocative of luxury and grandeur, giving the space a more lavish feel.
The resort reference added bright, open impressions and tropical colors, evoking a sense of extraordinary relaxation.
Meanwhile, the minimal design reference used a simple, uncluttered color palette and shape, creating a more refined interior.

In all three categories (artwork, car, and interior), features from the reference images were effectively transferred onto the source images.
Specifically, in the artwork category, color usage and unique brushwork from the artist were reflected; in the car category, forms and textures from animals were integrated; and in the interior category, spatial and atmospheric design elements were incorporated.
Consequently, the participants could accurately identify the reference pair used.
These findings suggest that presenting a concept as an image allows users to successfully reflect these features in another image.
All images used in the user study are presented in the Appendix.

%
%
\section{Limitations}
\label{sec:limitations}
This study had two main limitations.

First, our method relies on the assumption that features in the IP-Adapter's CLIP embedding space are disentangled.
If this assumption fails, ensuring accuracy becomes difficult.
Not all concepts represented by color, texture, shape, and other visual elements were fully disentangled in the $768 \times 4$-dimensional embedding space. 
Moreover, even if the embedding space were fully disentangled, the quality of the generated images would still depend on the representational capacity of the IP-Adapter's decoder (see the Appendix for failure outputs.)

Second, a challenge arose because the parameters ($\theta, d$) used to extract the features strongly affected the generated images.
Controllability is one of the strengths of our method. 
However, as with most generative models, it requires parameter tuning.

%
%
\section{Conclusion}
\label{sec:conclusion}
In this study, we propose \textit{Visual Concept Blending}, a method that effectively extracts and transfers shared or distinct features from multiple reference images onto a source image.
Using multiple reference images, we can distinguish which features of the reference images should be transferred, whereas existing VCT methods use only one reference image. 
By leveraging a partially disentangled embedding space in the IP-Adapter's CLIP, our approach can handle higher-level concepts such as shape transformation and motion.
While we acknowledge the limitation of disentanglement, the qualitative results showed diverse transformations across various domains, and our user study further confirmed that the participants could identify the intended reference images more accurately when these features were transferred.

In future studies, we plan to explore domain-specific training strategies and novel approaches for learning more disentangled embeddings.
Additionally, we aim to develop automatic parameter-setting techniques to further enhance the applicability and ease of use of Visual Concept Blending.
For example, computing embedding-level differences or a structural similarity measure (e.g., LPIPS, SSIM) between source and reference images, and adaptively adjusting $\theta$.
This will reduce trial-and-error, which can be time-consuming.

{
\small

}
\clearpage

\appendix

\onecolumn

\begin{center}
    {\Large \bfseries Supplementary Material for\\Zero-Shot Visual Concept Blending Without Text Guidance}
\end{center}

This supplementary material presents additional figures for the parameter sensitivity analyses (Sec.~3.2), qualitative results (Sec.~4.1), and images used in the user study (Sec.~4.2).
We also discuss failure cases of the generated outputs and ethical considerations.

\section{Additional Parameter Sensitivity Analyses}
\begin{figure}[!h]
  \centering
  \includegraphics[scale=0.2]{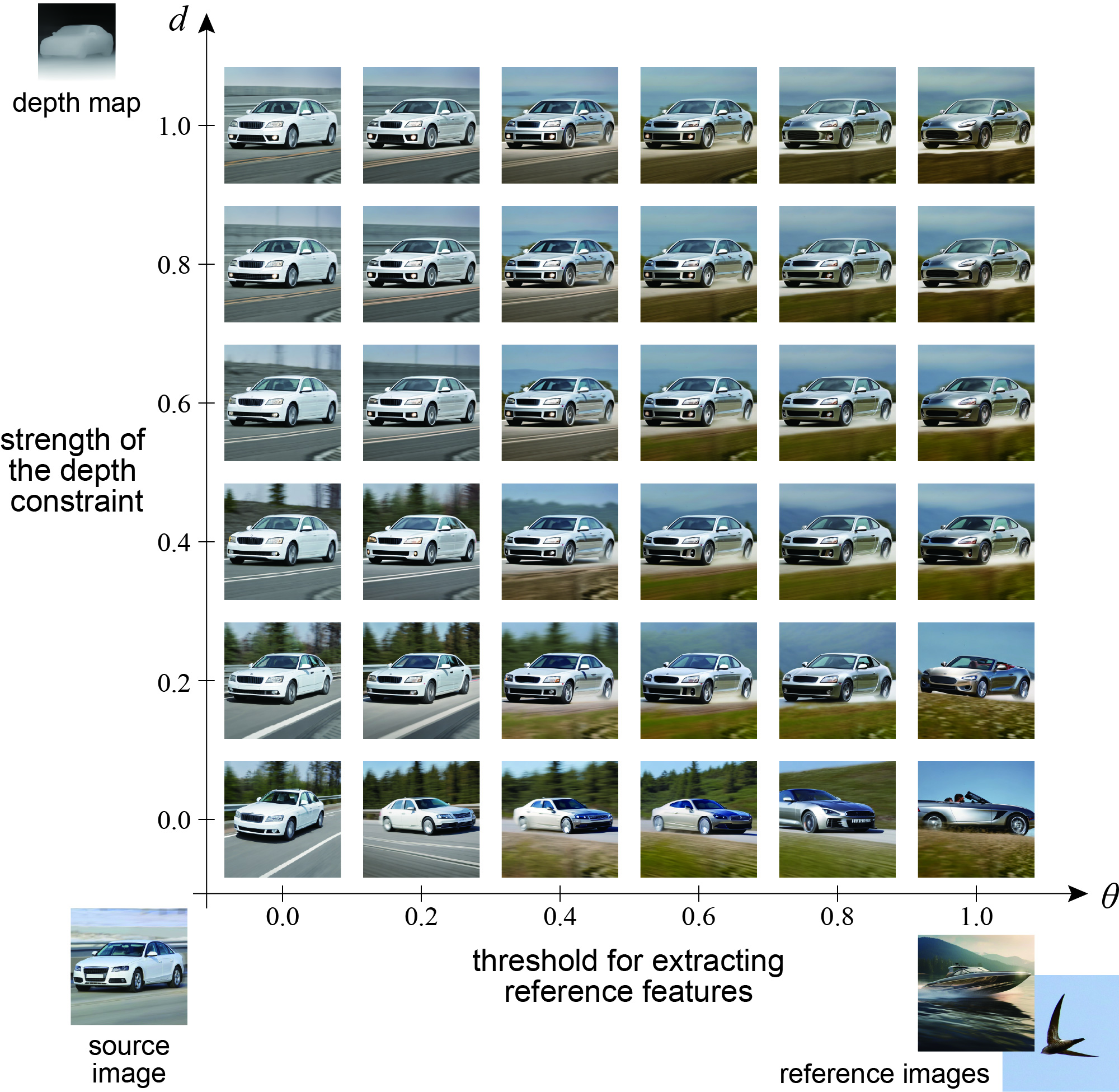}
  \caption{\textbf{Additional parameter sensitivity analysis 1.}}
  \label{batch2}
\end{figure}

Fig.~\ref{batch2} shows an additional parameter sensitivity analysis in which the reference images are a swift bird and a speedboat. 
Although these references share streamlined features, they belong to different domains. 
Therefore, even when the threshold is increased, the underlying shape of the car remains unchanged.

\begin{figure}[!h]
  \centering
  \includegraphics[scale=0.2]{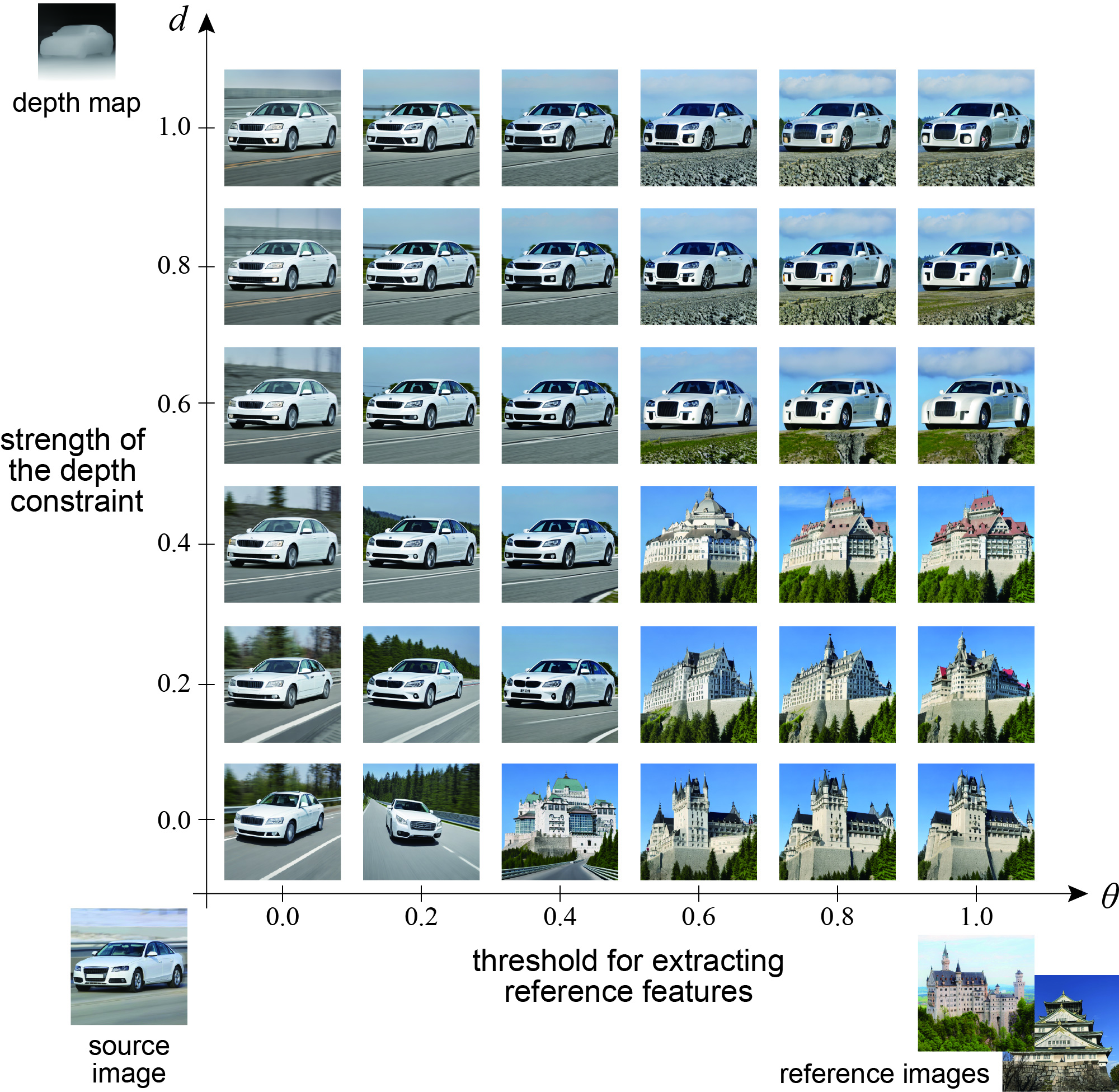}
  \caption{\textbf{Additional parameter sensitivity analysis 2.}}
  \label{batch3}
\end{figure}

\clearpage
Fig.~\ref{batch3} presents a similar example in which a German castle and a Japanese castle are used as reference images.
Both references convey a noble atmosphere and are characterized by tall, angular forms, which are clearly reflected in the generated images.
However, since both references belong to the same ``castle'' domain, increasing the threshold causes the original car concept to be overwritten by that of a castle.

\clearpage
\section{Additional Qualitative Results}
\begin{figure*}[!h]
  \centering
  \includegraphics[scale=0.2]{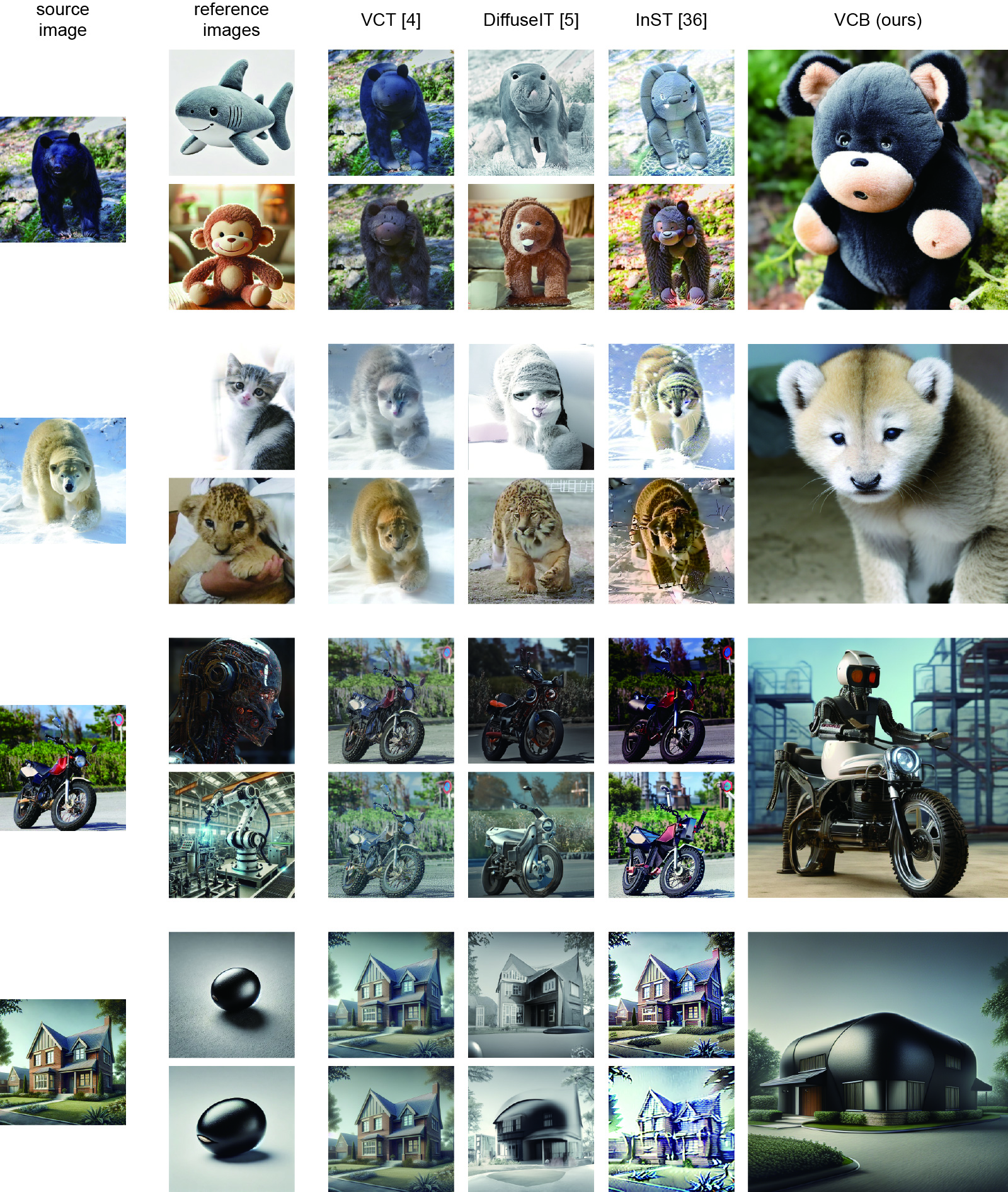}
  \caption{\textbf{Additional results on blending the common features of reference images.} For VCB, $\theta=0.4$ and $d=0.0$.}
  \label{qualitative_results_1_2}
\end{figure*}

\begin{figure*}[!h]
  \centering
  \includegraphics[scale=0.2]{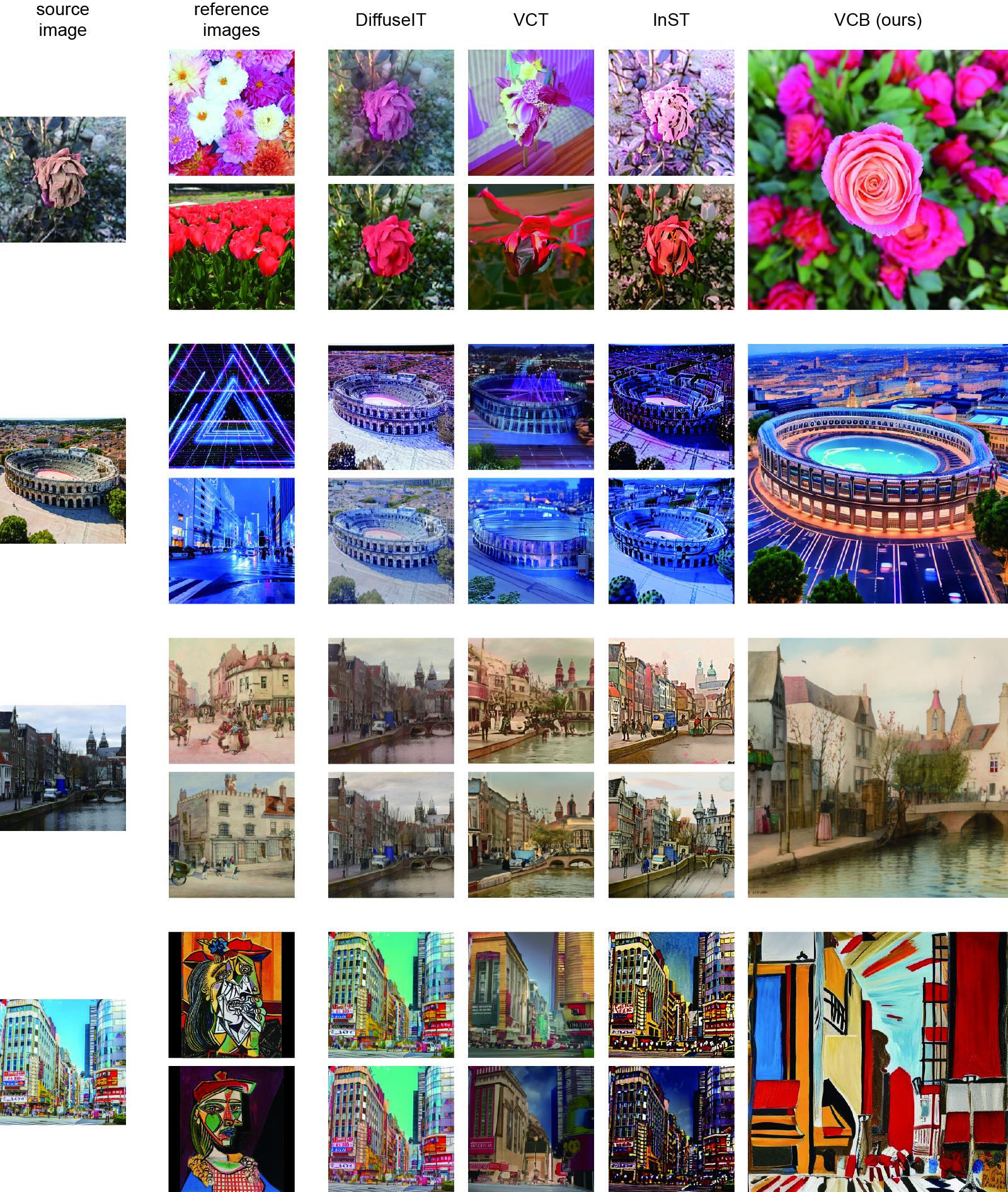}
  \caption{\textbf{Additional results on blending the common features of reference images.} For VCB, $\theta=0.4$ and $d=0.4, 1.0, 1.0, 1.0$, respectively.}
  \label{qualitative_results_1_3}
\end{figure*}

Figs.~\ref{qualitative_results_1_2} and \ref{qualitative_results_1_3} show additional qualitative results.
The proposed method successfully transfers concepts across a wide range of visual attributes.

\clearpage
\section{All Images Used in the User Study}
\begin{figure}[!h]
  \centering
  \includegraphics[scale=0.34]{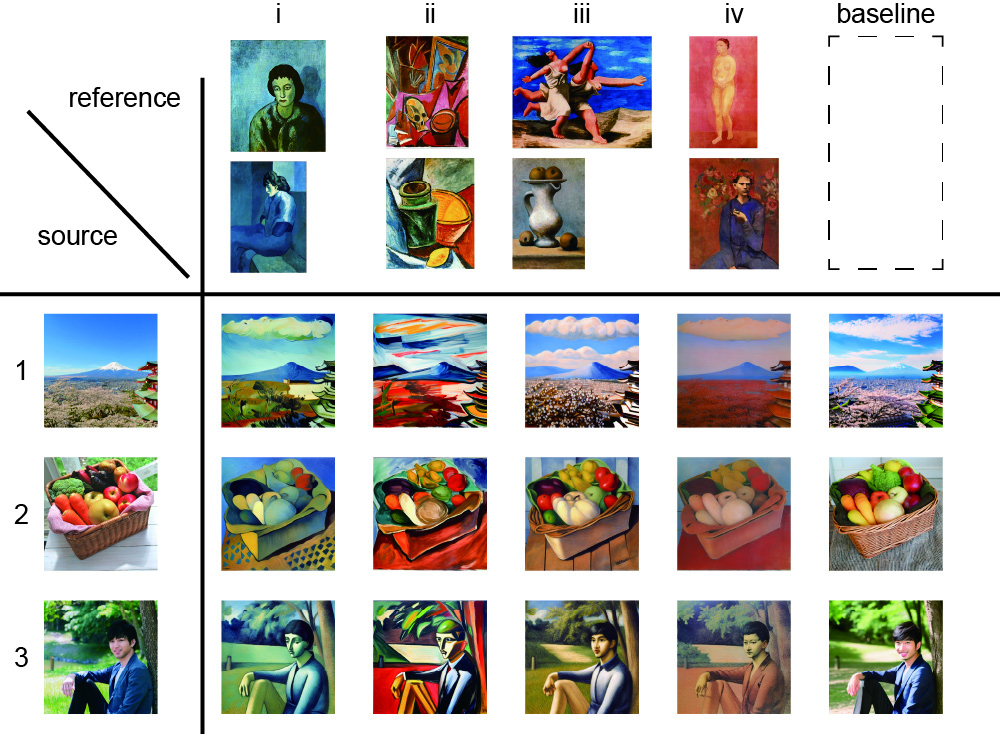}
  \caption{\textbf{All images used in the user study: (a) artwork.} $\theta=0.4, d=0.7$. Artwork references are from WikiArt (\url{https://www.wikiart.org/en/pablo-picasso/}).}
  \label{user_study_1_art}
\end{figure}

\begin{figure}[!h]
  \centering
  \includegraphics[scale=0.34]{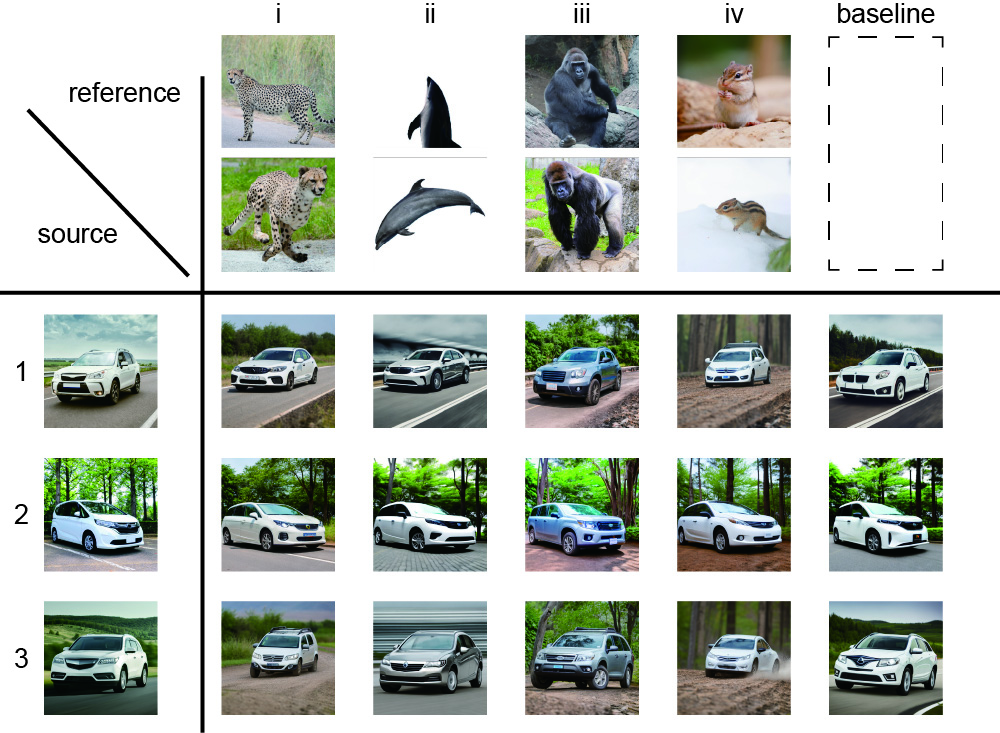}
  \caption{\textbf{All images used in the user study: (b) car.} $\theta=0.2, d=0.2$.}
  \label{user_study_1_car}
\end{figure}

\begin{figure}[!h]
  \centering
  \includegraphics[scale=0.34]{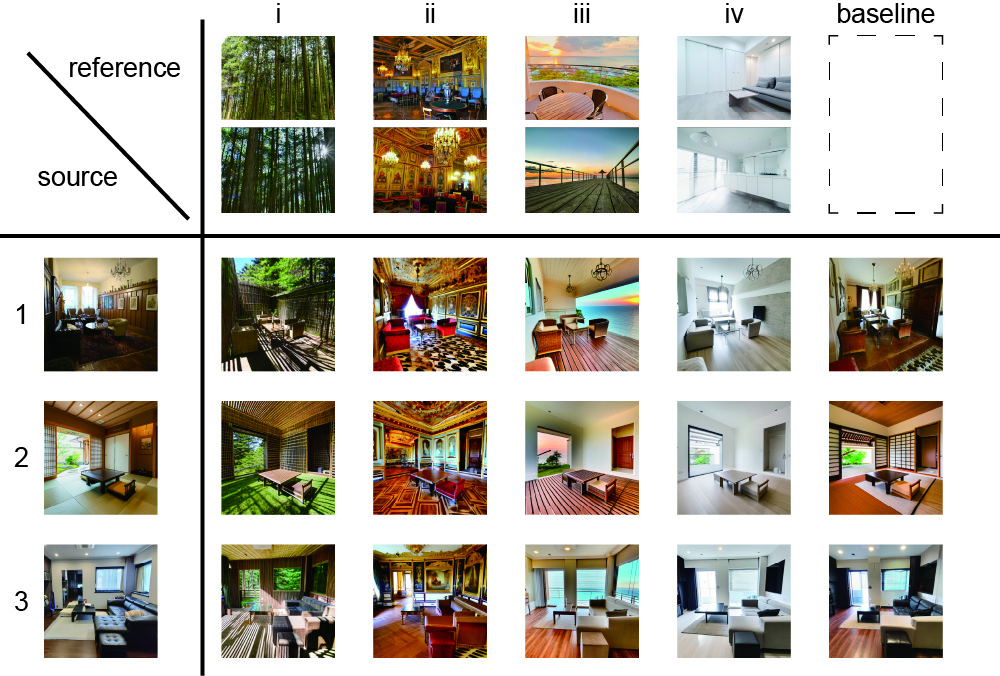}
  \caption{\textbf{All images used in the user study: (c) interior} $\theta=0.5, d=0.7$.}
  \label{user_study_1_interior}
\end{figure}

Figs.~\ref{user_study_1_art}, \ref{user_study_1_car}, and \ref{user_study_1_interior} display all images used in the user study.
For each pair of source and reference images, our method successfully transfers features in the intended direction, as confirmed by the user study.

\section{Failure Case Outputs}
\subsection{Case 1}
As mentioned in the Limitations section, the CLIP embedding of the IP-Adapter is not completely disentangled.
For example, refer to Fig.~\ref{user_study_1_art} (source 2 – reference ii).
In this output, certain characteristics of the reference image appear to be reflected, as confirmed in the user study.
However, Cubism is generally characterized by depicting subjects through geometric abstraction and expressing multiple perspectives simultaneously. 
The generated image does not clearly exhibit these defining features.
This suggests that the IP-Adapter's CLIP embedding struggled to effectively encode and transfer Cubist attributes to the source image.

\subsection{Case 2}
As also discussed in the Limitations section, the expressive capability of the decoder presents another challenge.
For instance, refer to Fig.~\ref{user_study_1_car} (source 3 – reference iv).
In this output, the model successfully integrates the features of a squirrel into a large vehicle, transforming it into a smaller and cuter form.
However, design flaws remain, notably the left rear wheel protruding unnaturally, rendering the generated result unsuitable for practical industrial applications.
One possible reason for the misplacement of the wheels in the generated image, despite their correct positioning in the source image, is the limited expressive capacity of the decoder.

\section{Ethical and Societal Impacts} 
In this work, we do not perform any model training; instead, we rely exclusively on a publicly available pre-trained model during inference.
Nevertheless, addressing ethical concerns such as privacy, consent, and copyright issues for both input and output images remains essential.
Regarding input images, users should ensure proper consent has been obtained from any depicted individuals and verify that reference images comply with applicable licenses or are in the public domain.
Even generated outputs may unintentionally incorporate copyright-protected characters or distinctive artistic elements, potentially inherited from the underlying pre-trained model or reference images.
To mitigate these risks, users are advised to carefully verify usage rights, obtain necessary permissions, and adhere strictly to the terms of service of the models employed.

Furthermore, our user study, conducted with Institutional Review Board approval, involved only non-sensitive or publicly available images for evaluating participants' recognition of transferred features.
No personally identifiable information was collected, and all participants provided informed consent prior to participating.

\end{document}